\documentclass[]{spie}  

\usepackage{amsmath,amsfonts,amssymb}
\usepackage{graphicx}

\usepackage[colorlinks=true, allcolors=blue, pdfborderstyle={/S/U/W 1}]{hyperref}

\usepackage{comment}
\newcommand{\mehmodified}[1]{{\color{black}#1}}
\newcommand{\modified}[1]{{\color{black}#1}}
\newcommand{\rpzmodified}[1]{{\color{black}#1}}
\newcommand{\newaddition}[1]{{\color{black}#1}}
\newcommand{\mmmodified}[1]{{\color{black}#1}}
\newcommand{\fsmodified}[1]{{\color{black}#1}}
\newcommand{\kjmodified}[1]{{\color{black}#1}}

\usepackage{multicol}
\usepackage{caption}
\title{\modified{Deep neural networks-based denoising models for CT imaging and their efficacy}
}

\author{Prabhat KC}
\author{Rongping Zeng}
\author{M. Mehdi Farhangi}
\author{Kyle J. Myers}

\affil{The Food and Drug Administration, Silver Spring, MD}

\authorinfo{Further author information: (Send correspondence to P. KC)\\P. KC E-mail: \href{mailto:prabhat.kc@fda.hhs.gov}{prabhat.kc@fda.hhs.gov}, Telephone: +1 (301) 796-9622\\Publication Citation: Prabhat KC, Rongping Zeng, M. Mehdi Farhangi, Kyle J. Myers, "Deep neural networks-based denoising models for CT imaging and their efficacy," Proc. SPIE 11595, Medical Imaging 2021: Physics of Medical Imaging, 115950H (15 February 2021); \url{https://doi.org/10.1117/12.2581418}.}

\begin{document} 
\maketitle

\begin{abstract}

Most of the Deep Neural Networks (DNNs) \mmmodified{based} CT image denoising literature shows \kjmodified{that DNNs} outperform traditional iterative methods \mmmodified{in terms of metrics} such as the RMSE, the PSNR and the SSIM. In many instances, using the same metrics, the DNN results from low-dose inputs are also shown to be comparable to their high-dose counterparts. However, these metrics do not reveal if the DNN results preserve the visibility of subtle lesions or if they alter the CT image properties such as the noise texture.

Accordingly, in this work, we seek to examine the image quality of the DNN results from a holistic viewpoint \kjmodified{for low-dose} CT image denoising. First, we build a library of \fsmodified{advanced} DNN denoising architectures. This library \kjmodified{is comprised of} denoising architectures such as the DnCNN, U-Net, Red-Net, GAN, etc. Next, each network is modeled, as well as trained, such that it yields its best performance \mehmodified{in terms} of the PSNR and SSIM. As such, data inputs (e.g. training patch-size, reconstruction kernel) and \mehmodified{numeric-optimizer inputs (e.g. minibatch size, learning rate, loss function) are accordingly tuned.} Finally, outputs from thus trained networks are further subjected to a series of CT bench testing metrics such as the \rpzmodified{contrast-dependent MTF, the NPS and the HU accuracy}. These metrics are employed to perform a more nuanced study of the resolution of the DNN outputs’ \kjmodified{low-contrast features, their noise textures, and their CT number accuracy to better understand the impact each DNN algorithm has on these underlying attributes of image quality}.

\end{abstract}


\section{INTRODUCTION}
\label{sec:intro}  

Computed Tomography (CT) based imaging modalities are extensively used to perform non-evasive medical \fsmodified{diagnoses} and to \kjmodified{make treatment decisions}. At present, it is \kjmodified{estimated} that about 70 million scans are performed per year in the U.S. alone \cite{ct_per_year_us}. The same number was about 50 million scans in \kjmodified{the early} 2000s and about 12 million in \kjmodified{the early} 1990s \cite{ct_increasing_trend}. This ever-increasing trend in the use of CT imaging has brought about a growing concern \kjmodified{regarding the} risks associated with radiation exposure to patients \cite{seeram2015computed}. Accordingly, the CT imaging community has been persistently working under the guiding principle of as low as reasonably achievable (ALARA) to lower the CT dose without compromising the images' diagnostic information\cite{ehman_img_quality_def,chen_redcnn}. All research endeavors performed to realize the ALARA goal can be broadly classified into two branches. They include hardware upgrades and advancement in reconstruction algorithms. In this paper, we will focus on the \fsmodified{latter} part.

Since the inception of CT imaging in 1970s, the filtered back-projection (FBP) algorithm has been the predominant reconstruction module in \kjmodified{almost all} CT-scanners \cite{hendee_med_ct_book}. However, the past decade saw a decent uptake \kjmodified{in use of iterative reconstruction (IR) methods in new} generation scanners. For instance, the model-based IR (MBIR) \cite{mbir_in_ct_scanner} has been incorporated in several of the new scanners with the aim \kjmodified{of obtaining} high quality CT images with \kjmodified{low-dose} as per the ALARA philosophy. However, the long computation time of the MBIR technique may hinder its usage in a clinic setting when instant image outputs are required. Consequently, researchers have begun to \kjmodified{explore use of deep learning (DL) \cite{first_dnn_2_denoise} methods for CT image denoising} \cite{kim_vgg_paper} to overcome the computational time limit. The DL methods gained prominence due to their unprecedented success in image classification and solving various computer vision related problems. \fsmodified{Hoping for a similar success in the domain of CT image denoising, there has been a significant uptake on research projects that focus in use of DL methods to enhance low-dose acquisitions. Usually, these endeavors show that the DL results supersede the IR results on the basis of global fidelity or noise metrics such as the Structural Similarity Index (SSIM), the Root-Mean-Square Error (RMSE) and the Peak Signal-to-noise ratio (PSNR) that are appropriate for computer vision related tasks}. These metrics do not provide answer to the fundamental question on how the \kjmodified{DL-based} denoised solution performs as compared to the FBP results - from the normal dose CT (NDCT) - on local structures. The answer to this question is of utmost importance to enable doctors to make accurate clinical decisions. Hence, in this work, we seek to \kjmodified{analyze} the image quality of denoised results from multiple networks with the help of the standard CT image evaluation metrics \newaddition{(or the CT bench tests) such as the Modulation Transfer Function (MTF), the Noise Power Spectrum (NPS), and \rpzmodified{the HU accuracy}.}

 To sum up, our unified approach to determine the efficacy of DL frameworks \fsmodified{consists of}, first, denoising \fsmodified{the low-dose CT (LDCT)} images (at $25\%$ dose level) by independently optimizing these frameworks to yield their respective best PSNR and SSIM values on \newaddition{a tuning dataset}. Subsequently, we employ qualitative, as well as quantitative CT bench testing tools, \kjmodified{including the NPS and MTF, \fsmodified{to report} an overall analysis on these building blocks of} image quality from a given DNN.

\section{method}\label{sec:method}
A typical neural networks-based CT image denoising framework seeks to learn a function, $f$, that maps a LDCT image to its corresponding NDCT image \cite{wgan_paper}. The function, $f$, is parameterized with network weights, $\theta$. These weights are estimated from a training set by solving the following objective function:
\begin{equation}
\hat{\theta}\leftarrow\underset{\theta}{\text{arg min}\ }\ell(f(\mathbf{X};\theta),\mathbf{Y})\, ,\label{eq:general_loss_func} \tag{M}
\end{equation}
where $\ell(\cdot)$ denotes a loss function and $\mathbf{X},\mathbf{Y}\thinspace\in\thinspace\mathbf{R}^{m\times n}$ represent, complimentary, LDCT and NDCT images from the training set.   
We make use of the Low-dose Grand Challenge (LDGC) dataset \cite{ldct_data}, shared to us by the Mayo Clinic, to train $\theta$ in model (\ref{eq:general_loss_func}). The training set \fsmodified{comprises $1553$} images of size $512 \times 512$ obtained from $6$ different patients. Likewise, our validation/tuning set consists of $40$ images randomly pre-selected from the same $6$ patients before formulating the training set. Finally, our test set \fsmodified{comprises $223$} images obtained from a seventh patient. Whilst training and validation sets play \kjmodified{a crucial role in estimating} $\hat{\theta}$, there are a number of other important \newaddition{operations/options/choices} that also affect how well a DL denoiser is trained and, ultimately, the image quality of its denoised outputs. We have broadly classified these choices under the following headings:
\begin{enumerate}
  \item[{1)}]Data and feature pre-processing: We check to see if networks trained using CT images \kjmodified{obtained} from different reconstruction kernels (i.e.\thinspace \thinspace sharp vs.\thinspace \thinspace smooth filters) yield different performance. Additional \mehmodified{feature pre-processing} choices such as data augmentation, patch-size and data normalization are also thoroughly investigated to achieve the best network performances. Below is an overview of these choices considered in this study:
  \begin{itemize}
      \item Scaling: Each image-pair is downscaled by factors $0.6$ and $0.8$.
      \item Rotation: Each image-pair is rotated by either $90^{\circ}\ \text{or }180^{\circ}\ \text{or } 270^{\circ}$ and flipped either left-right or up-down.
      \item Dose: Each image-pair is blended by calculating an additional LDCT image as \\ NDCT $+$ $\gamma(\text{LDCT}_{\text{quater dose}}-\text{NDCT})$, where $\gamma\ \epsilon\ \mathcal{U}[0.5,1.2]$ with $\mathcal{U}$ denoting the uniform distribution. 
      \item \fsmodified{Normalization types: Separate learnings with the training set normalized to unity range $[0, 1]$ in one instance and re-scaled to yield non-negative values in another instance.
      \item Patch-sizes: Separate learnings with the training set patched to each of the following sizes - $32\times32,\ 55\times55,\ 64\times64,\text{ and }96\times96$.}
  \end{itemize}
  \item[{2)}] Neural Networks: Different DL architectures have different denoising capabilities. Here, we begin from a simple $3$-layered convolutional neural network (CNN3). Subsequently, we move to optimize denoising performances of high-end DL networks such as, the feed-forward denoising CNN (DnCNN) \cite{zhang_DnCNN}, the Residual Encoder-Decoder CNN (REDCNN) \cite{red_net_orig,chen_redcnn}, the Dilated U-shaped DnCNN (DU-DnCNN) \cite{unet_git}, and the Generative Adversarial Networks (GAN) \cite{goodfellow_GAN}. \newaddition{In this study, the DnCNN is $17$-layered, the REDCNN is $10$-layered with $5$ encoding convolution layers and $5$ decoding deconvolution layers, the DU-DnCNN is $10$-layered, and the GAN \fsmodified{comprises $20$-layered} ResNet \cite{srgan_paper} as its generator network and the 10-layered DnCNN as its discriminator network. For more information on the architectures, readers are suggested to refer to their respective citations.}
  \item[{3)}]Computational Optimization: We incorporate data distributed based network training on multi-GPUs using Horovod \cite{horovod_origi_paper} within PyTorch framework \cite{pytorch_package}. \modified{Accordingly, we adhere to the results reported by Goyal et al.\cite{horovod_lr}} to efficiently train our networks in terms of minibatch sizes and learning rates. \fsmodified{The minibatch sizes considered for this study include $64,\ 128,\ 256,\ 512$. The learning rates considered for this study include $10^{-1},\ 10^{-2},\ 10^{-3},\ 10^{-4}$.} Likewise, we also analyze several loss functions to, ultimately, \modified{determine the best performing} one for each of the aforementioned networks for our denoising task. These loss functions are listed below:
   \begin{multicols}{2}
   \begin{enumerate}
    \item[{(A1)}] $\frac{1}{m}\underset{i=1}{\overset{m}{\sum}}||f(\mathbf{X}^{i};\theta)-\mathbf{Y}^{i}||^{2},$ 
    \item[{(A3)}]$ \frac{1}{m}\underset{i=1}{\overset{m}{\sum}}|f(\mathbf{X}^{i};\theta)-\mathbf{Y}^{i}|,$
    \item[{(A5)}] $\frac{1}{m}\underset{i=1}{\overset{m}{\sum}}||f(\mathbf{X}^{i};\theta)-\mathbf{Y}^{i}||^{2}+\frac{\beta}{2}||\theta||^{2},$
    \item[{(A2)}]$\frac{1}{m}\underset{i=1}{\overset{m}{\sum}}||f(\mathbf{X}^{i};\theta)-\mathbf{Y}^{i}||^{2}+\frac{\lambda}{2}|f(\mathbf{X};\theta)|,$
    \item[{(A4)}]$ \frac{1}{m}\underset{i=1}{\overset{m}{\sum}}||f(\mathbf{X}^{i};\theta)-\mathbf{Y}^{i}||^{2}+\frac{\lambda}{2}|\nabla f(\mathbf{X};\theta)|,$ 
    \item[{(A6)}]$\alpha(1-\text{MS-SSIM}[f(\mathbf{X};\theta), \mathbf{Y}])+(1-\alpha)\cdot{|f(\mathbf{X};\theta)-\mathbf{Y}|},$
    \end{enumerate}
    \end{multicols}
    \begin{itemize}
        \centering
        \item[{(A7)}]\rpzmodified{$\underset{G}{\text{min}}\ \underset{D}{\text{max}}\left(\mathbb{E}_{x\sim p_{\text{data}}(x)}[\log D(x)]+\mathbb{E}_{z\sim p_{z}(z)}[\log(1-D(G(z)))]\right).$}
    \end{itemize}
\end{enumerate}

Loss functions in (A1), (A3), and (A5) represent the Mean-Squared Error (MSE), the Mean Absolute Error (MAE), and MSE with the weight decay term (MSE+wd) respectively. Similarly, the loss function in (A2) encompasses MSE with the L$_{1}$ image prior (MSE+L$_{1}$) and the one in (A4) includes MSE with the total variation (TV)-image prior (MSE+TV). $\beta$ in (A5) and $\lambda$s in (A2), (A4) are heuristically determined \kjmodified{for each} given network. (A6) incorporates the Multi Scale–SSIM (MS-SSIM) loss\cite{ms_ssim_ct} with MAE between network output and its corresponding target. We refer to (A6) as MS-SSIM+L$_{1}$ and its $\alpha$ is empirically set as 0.84 \cite{ms_ssim_cv}. \newaddition{Finally, (A7) represents the adversarial min-max loss function for the GAN with $D$ as the discriminator network and $G$ as the generator network.}

\newaddition{We make use of the aforementioned method to estimate $\hat{\theta}$ for each network in two different manners. These two approaches are listed below:

\begin{enumerate}
    \item[{(i)}]Learning to optimize/tune global metrics: In this paradigm, all the networks with their corresponding choices (points $1$ through $3$) are optimized – using the tuning set – to yield the best values w.r.t the PSNR and SSIM (akin to several \kjmodified{DL-based} CT denoising publications). Then we report their performances in terms of their resolving capacities and their noise textures. In particular, we perform contrast-dependent MTF analysis on a simulated CATPHAN$600$ CT image at four different levels, namely $900,\ 340,\ 120,\ \text{and}\ -35$ Hounsfield Units (HU). Likewise, the NPS test is performed on 50 noisy CT simulations of a reconstructed cylindrical water phantom.
    
    \item[{(ii)}]Learning to optimize/tune CT bench tests: In this DL approach, all the networks are tuned to give an overall best performance w.r.t the MTF value, noise texture, and \rpzmodified{HU accuracy}. Even in this approach we continue to keep track of the PSNR and SSIM values so that the network learning does not sway away in a manner that the learnt weights overfit CT bench tests and underperform on the tuning set comprising of \rpzmodified{patients' CT images}. Nonetheless, here we favor the network choices that yield an overall better result in terms of the CT bench tests albeit with \kjmodified{reduced} PSNR, SSIM values over the choices that give the best PSNR, SSIM values with \kjmodified{decreased performances} on the CT bench tests.  
\end{enumerate}
}

\section{RESULTS}\label{sec:results}
\subsection{From global metrics based tuning}\label{sec:glob_result}
\begin{figure}[h]
\centering
\captionof{table}{PSNR and SSIM values (i) for different setup of the CNN3 architecture and (ii) from best tuned algorithms}
\includegraphics[width=1\linewidth]{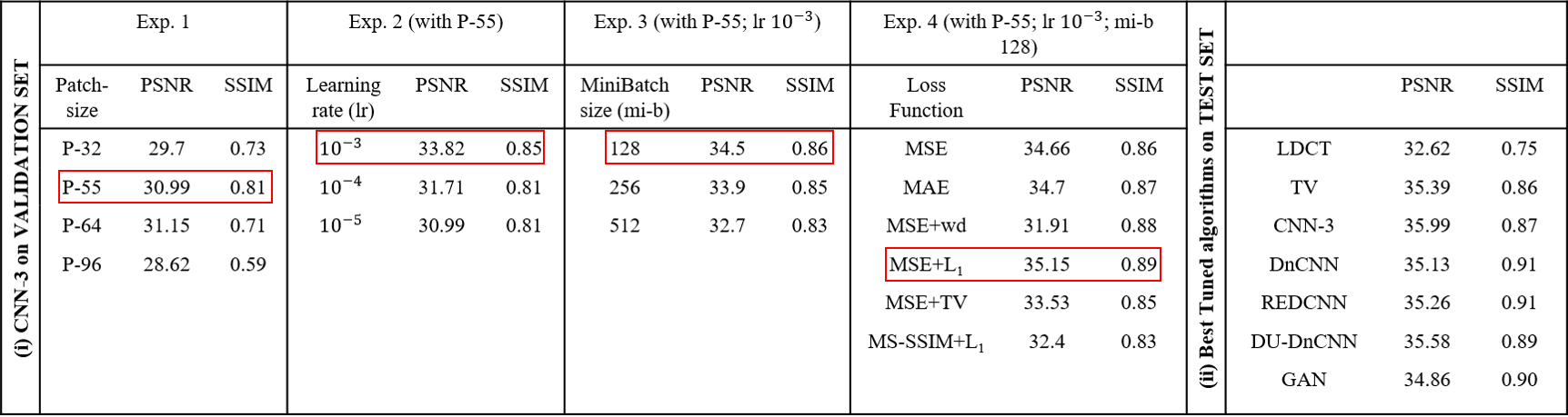}
\label{img:Table1}
\end{figure}

We adhere to the results obtained by Zeng et al.\cite{rpz_generalizability} for some of the data pre-processing choices such as, slice thickness and reconstruction kernel. They report that the DNNs (in particular REDCNN) trained on CT images with \kjmodified{a slice thickness of $3$ mm reconstructed using a sharp kernel} outperform those trained with CT images of $1$ mm thickness reconstructed a using smooth filter kernel from the MTF viewpoint. As such, we incorporate LDCT-NDCT pairs from the LDGC repository with 3 mm thickness that are reconstructed employing sharp filter for all of our supervised training experiments. Second, all five networks \kjmodified{resulted in} their respective best performances - in terms of the global metrics based values i.e.$\thinspace$ the PSNR, SSIM, RMSE - for \textit{unity} normalization with the training set normalized to the range $[0, 1]$. Third, we did not observe any gain in performance w.r.t the global metrics for any of the networks with the data augmentation from strategies like rotation and scaling. With these pre-processing options set, we proceed to \kjmodified{investigate} other factors affecting a DL framework’s performance as described in section \ref{sec:method} (points $1$-$3$). A summary of experiments for one of the networks, i.e. CNN3, is listed in \mmmodified{Table} \ref{img:Table1}.

Table \ref{img:Table1}(i) demonstrates that patch size $55\times55$ (P-$55$) yields the best performance w.r.t the PSNR and SSIM values for the CNN3 in experiment (exp.)$\ 1$. The option that yields the best result in the exp.$\thinspace1$ is forwarded into exp.\thinspace$2$ and that from exp.$\thinspace2$ into exp.$\thinspace3$ and so on. Accordingly, we see that the CNN3 achieves its highest global metric performance on the tuning set with P-$55$; learning rate (lr): $10^{-3}$; mini-batch size (mi-b): $128$; and MSE+L$_{1}$ set up. Similar analyses for the rest of the networks \fsmodified{yield}:
DnCNN: (P-$55\ |$ lr:$10^{-4}\ |$ mi-b:$32\ |$ MSE+L$_{1}\ |\ \lambda$:$10^{-7}$), REDCNN: (P-$96\ |$ lr:$10^{-4}\ |$ mi-b:$32\ |$ MSE), DU-DnCNN: (P-$55\ |$ lr:$10^{-3}\ |$ mi-b:$64\ |$ MSE+L$_{1}\ |\ \lambda$:$10^{-7}$), GAN: (P-$55\ |$ lr:$10^{-4}\ |$ mi-b:$64$). Thus tuned DNNs are \kjmodified{applied to} the test set and the resulting PSNR and SSIM values, along with the ones from the Total Variation (TV) based IR method, are listed in Table 1(ii).
\begin{figure}[h]
\centering
\includegraphics[width=0.9\linewidth]{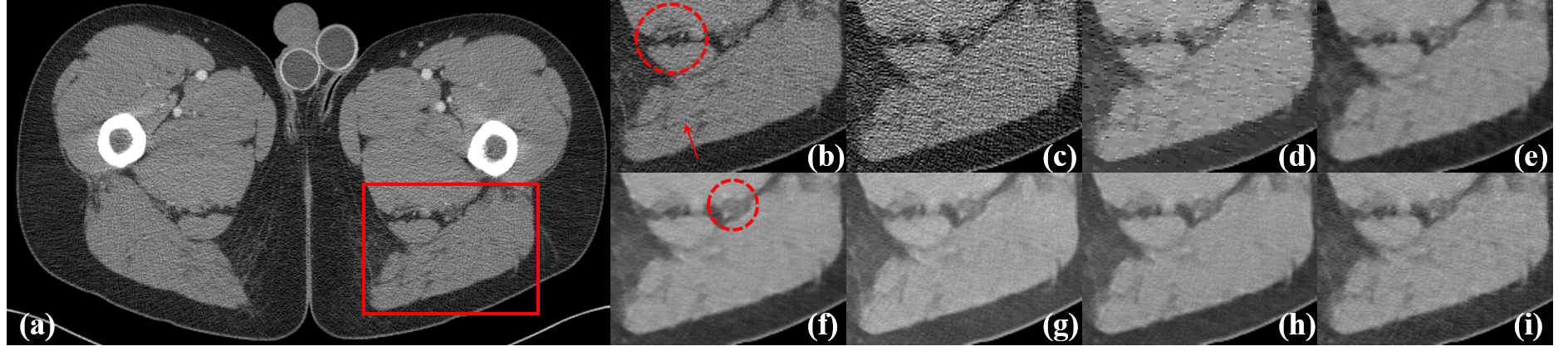}
\caption{(a) Full sized NDCT image. Zoomed view of the ROI of the (b) NDCT [PSNR$\ |\ $SSIM]  and its (c) LDCT [$34.94\ |\ 0.81$] counterpart. Denoised results of the previous LDCT image from (d) TV[$36.65\ |\ 0.91$] (e) CNN3 [$38.19\ |\ 0.89$] (f) RED-CNN [$36.79\ |\ 0.94$] (g) GAN [$33.74\ |\ 0.89$] (h) DnCNN [$36.36\ |\ 0.92$] (i) DU-DnCNN [$36.76\ |\ 0.89$]. The display window consists of (W:$491$ L:$62$) HU.}
\label{img:glob_abdomin_crop_view}
\end{figure}

Outputs from the trained DNNs for one of the test image are depicted in figs.$\ $\ref{img:glob_abdomin_crop_view}(e-i). They represent denoised results of a LDCT image shown in fig$.\ $\ref{img:glob_abdomin_crop_view}(c). The gain in the PSNR and SSIM values of the DNN outputs (figs.$\ 1$(e-i)) is also reflected in their reduced noise levels in comparison to their LDCT counterpart (fig$.\ $\ref{img:glob_abdomin_crop_view}(c)) from the visual standpoint. However, low-contrast features which are vividly clear in the NDCT image, such as the \textit{M}-shaped feature in fig.$\ $\ref{img:glob_abdomin_crop_view}(b) (indicated by upward arrow), appear mostly diminished in all of the DL-based solutions (i.e. figs.$\ $\ref{img:glob_abdomin_crop_view}(e-i)). Likewise, the anatomical features inside the dotted circle in the NDCT image (fig.$\ $\ref{img:glob_abdomin_crop_view}b) that are easily discernable appear in a coalesced form in all DL results (figs.$ $\ref{img:glob_abdomin_crop_view}(e-i)). Finally, all the DL results encompass high density \kjmodified{dot-like} features (as indicated by a circle in fig.$\ $\ref{img:glob_abdomin_crop_view}(f)) that are absent in the NDCT image (fig.$\ $\ref{img:glob_abdomin_crop_view}(b)).     

\begin{figure}[h]
\centering
\includegraphics[width=0.70\linewidth]{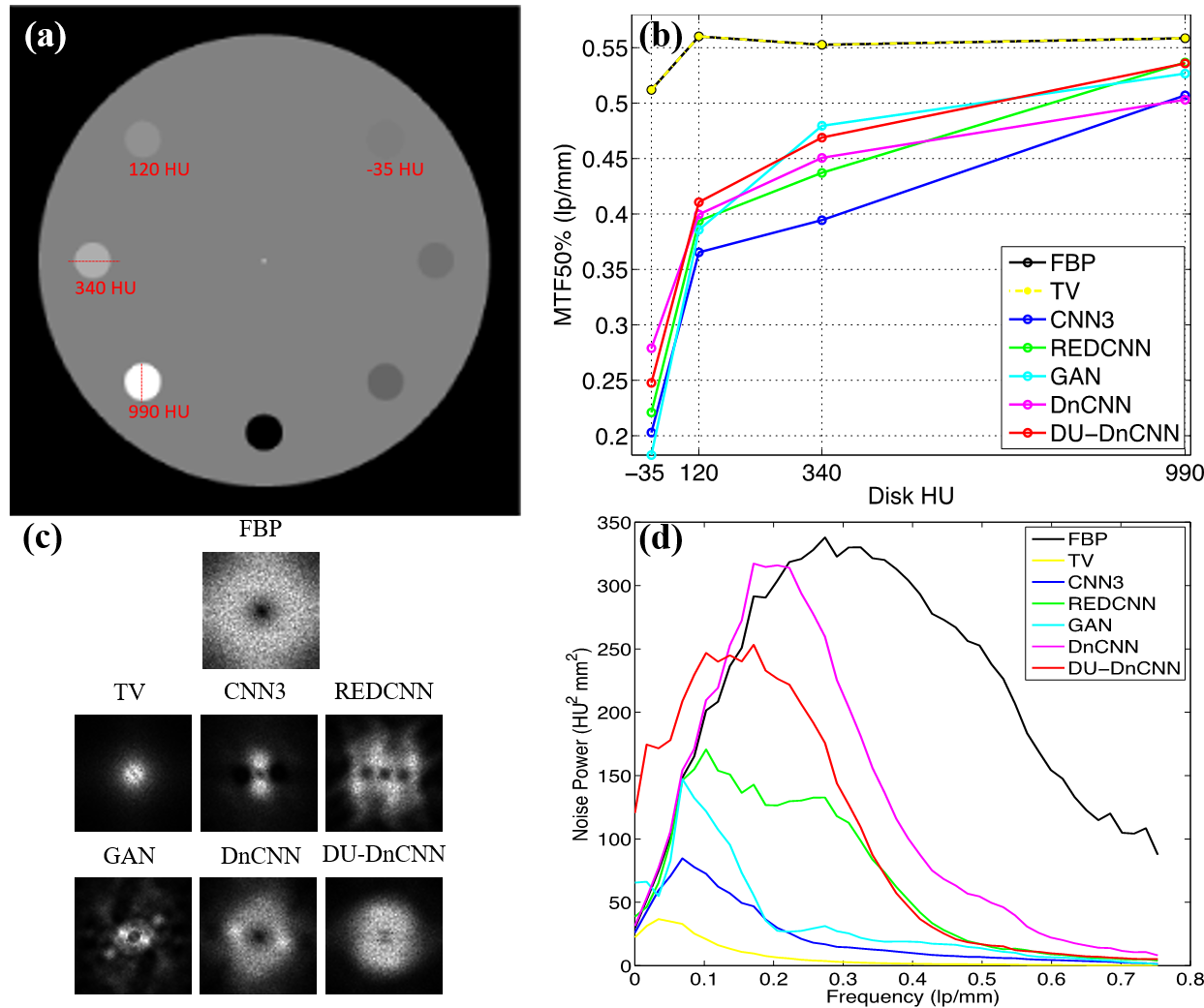}
\caption{(a) Simulated phantom that mimics contrast levels in the CATPHAN$600$. (b) MTF$50\%$ plot of the networks applied on the CATPHAN$600$ reconstructed using the FBP sharp kernel. (c) $2$D NPS images and (d) radial $1$D NPS curves of the networks applied on the noisy realizations of the reconstructed cylindrical phantom}
\label{img:nps_mtf_4rm_glob}
\end{figure}

The inability of the DNNs to accurately resolve small features can be readily explained with the MTF and NPS plots in figs.$\ $\ref{img:nps_mtf_4rm_glob}(b,d). We see that MTF$50\%$ values for all the DNNs are substantially \fsmodified{lower relative} to that of the FBP-sharp filter for low-contrast disks such as $120$ HU and $-35$ HU. Additionally, the radial $1$D NPS plot in fig.$\ $\ref{img:nps_mtf_4rm_glob}(d) reveals that the lower-mid to high-frequency bands (i.e. above $0.4$ lp/mm) are suppressed by all of the DNNs. \kjmodified{The high-frequency smoothing characteristic of the DNNs explains the blurring that is visually evident in figs.$\ $\ref{img:glob_abdomin_crop_view}(e-i). Also, the blurring observation is consistent with the MTF plot in the fig.$\ $\ref{img:nps_mtf_4rm_glob}(b).}

\subsection{From CT bench tests based tuning}\label{sec:ct_bench_result}

\subsubsection{HU Accuracy}\label{sec:ct_numbers}
\begin{figure}[h]
\centering
\includegraphics[width=0.60\linewidth]{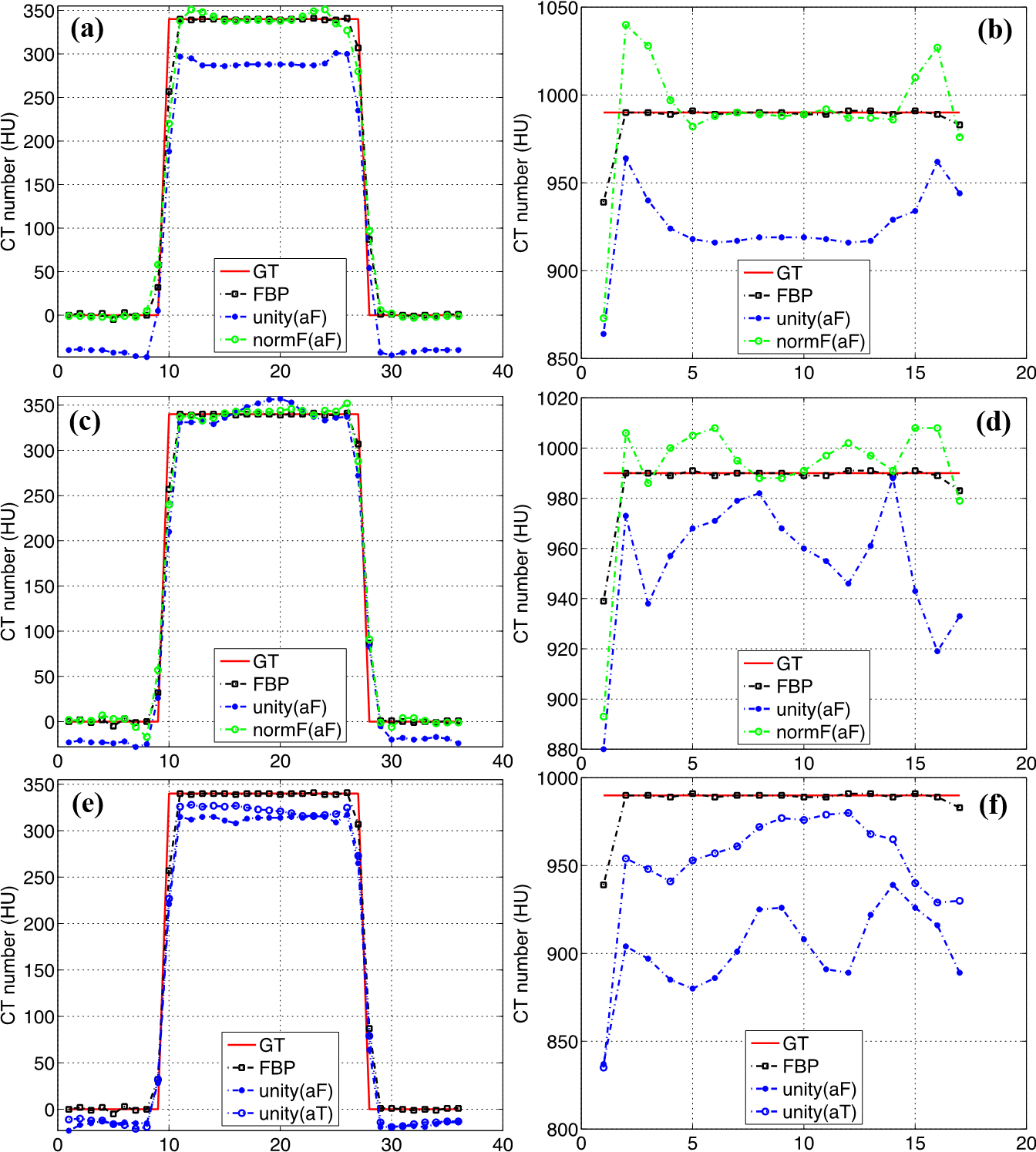}
\caption{Line-plot analysis along the dotted line of the $340$ HU disk in the fig.$\ $\ref{img:nps_mtf_4rm_glob}(a) from (a) CNN3 (c) REDCNN (e) GAN trained making use of different normalization types and data augmentation forms. A similar line-plot analysis for the $990$ HU disk from (b) CNN3 (d) REDCNN (f) GAN.}
\label{img:hu_no_norm}
\end{figure}

\begin{figure}[h]
\centering
\includegraphics[width=1.00\linewidth]{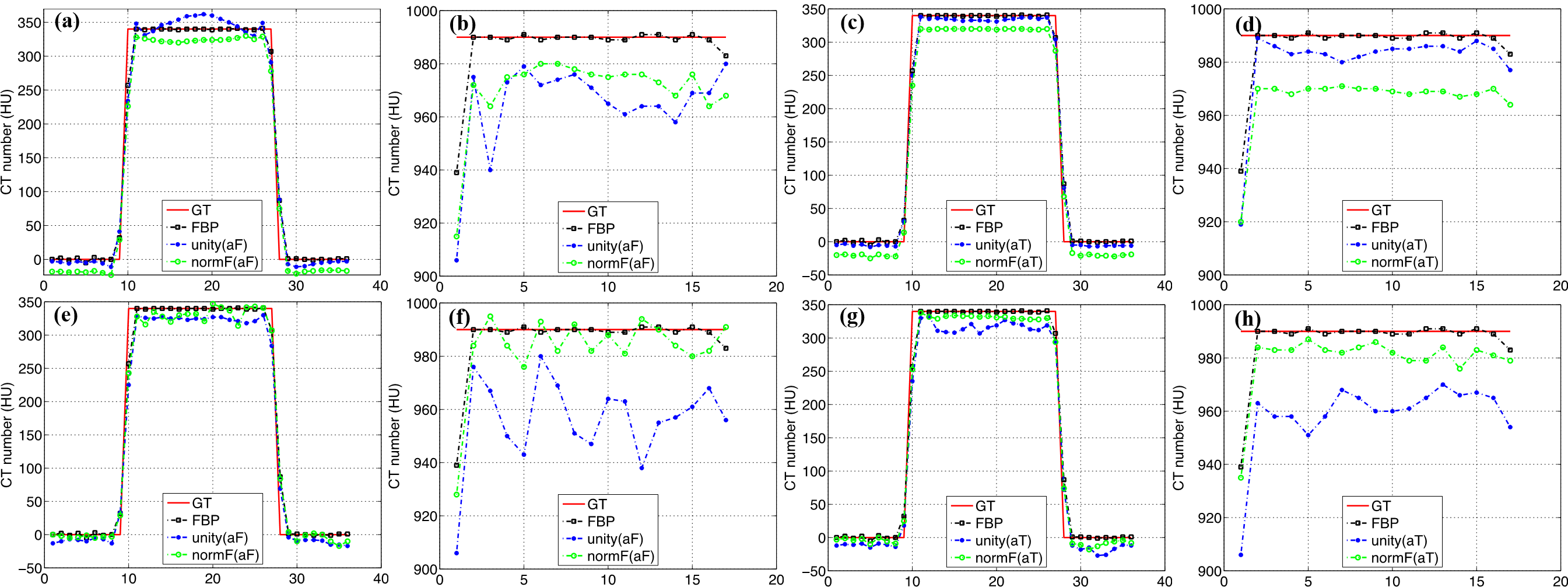}
\caption{Line-plot analysis along the dotted lines of the $340$ HU and $990$ HU disks in the fig.$\ $\ref{img:nps_mtf_4rm_glob}(a) from the DnCNN in (a) through (d) and from the DU-DnCNN in (e) through (f) trained making use of different normalization types and data augmentation forms.}
\label{img:hu_uni_norm}
\end{figure}

First, we \kjmodified{found that the image intensity values} of DL-based solutions were inconsistent across the test set via visual inspections. Accordingly, we make use of the CATPHAN$600$ to gain an insight on the performances of these networks \newaddition{in terms of the HU accuracy} \cite{ct_number_on_phan, ct_number_on_images}. More specifically, we perform line-plot analysis along the dotted red lines of the CATPHAN$600$ depicted in the fig.$\ $\ref{img:nps_mtf_4rm_glob}(a). After some heuristic-based experiments we realize that the nature of normalization type \rpzmodified{(employed on the training set)} and the presence of augmented data (during the learning phase) play key roles in ensuring HU accuracies of the learnt networks in par with that from the NDCT. A simple depiction is provided in figs.$\ $(\ref{img:hu_no_norm}-\ref{img:hu_uni_norm}). These figures illustrate line-plots along two differing contrasts i.e. $340$ HU and $990$ HU. In the plots, GT and FBP refer to the lines resulting from the ground truth and filtered backprojection based reconstruction of the CATPHAN$600$. As stated earlier, \textit{unity} denotes a deep learning implementation where its training dataset has been normalized to the range $[0, 1]$ and \textit{normF} refers to a learning where its corresponding training dataset has not been normalized and, rather, rescaled to exhibit non-negative values by (simply) adding $1024$ HU. Finally, \textit{aT}, represents a training set that has been subjected to down-sampling, rotation and dose-based augmentations, as explained in section \ref{sec:method}, while \textit{aF} denotes the one that does not exhibit any forms of augmentation. \newaddition{Note that, all the other network choices (exp. $1$–$4$, table \ref{img:Table1}) are still made to yield the best global metric values on the tuning set.}

From the perspective of the normalization, it is clear from figs.$\ $\ref{img:hu_no_norm}(a-b) that the CNN3 yields a higher HU accuracy from a learning whose training data exhibits \textit{normF} than the one with \textit{unity}. The same is the case with the REDCNN as illustrated in figs.$\ $\ref{img:hu_no_norm}(c-d). For training of the GAN, we only make use of the \textit{unity}-based normalization as we adhere \kjmodified{to its} mathematical assumption i.e.$\ 0$ is used to denote fake data and $1$ to denote true data\cite{goodfellow_GAN}. Still, note that the GAN trained with the augmented data i.e \textit{aT} (fig.$\ $\ref{img:hu_no_norm}(d)) achieves higher HU accuracy than the one trained without augmentation i.e. \textit{aF} (fig.$\ $\ref{img:hu_no_norm}(e)).

For the non-augmented (\textit{aF}) training case of the DnCNN, it is not clear as to which one of the two, the \textit{unity} or \textit{normF} based normalization, achieves higher accuracy when both of the contrasts,  $340$ HU and $990$ HU, are considered as depicted in figs.$\ $\ref{img:hu_uni_norm}(a-b). Yet, figs.$\ $\ref{img:hu_uni_norm}(c-d) illustrate that the \textit{unity}-based learning with augmentation aids the network to achieve higher HU accuracy that the one learnt with the \textit{normF} for the two differing contrasts. A similar analysis for the DU-DnCNN reveals that a learning from the augmented and \textit{normF} based training data set attains a higher degree of HU accuracy than a \textit{unity} based learning (figs.$\ $\ref{img:hu_uni_norm}(e-h)).

Even after performing all these CT numbers-based analyses, it is still wise to keep in mind that certain models have an inherent tendency to perform better \kjmodified{than others} for CT bench testing on \rpzmodified{standard phantoms} like the CATPHAN$600$. For instance, the TV-based denoiser will, almost, invariably outperform any other denoisers for a line-plot analysis on the CATPHAN$600$ due to its piece-wise prior term. Thus, we proceed \kjmodified{with caution} and flexibility to re-evalute/change our current pre-processing choices based on additional findings from other CT bench testing methods. For now, we move forward to the next subsection with the augmented-unity based pre-processing for the GAN, DnCNN architectures; and the augmented-normF based pre-processing for the CNN3, REDCNN, and DU-DnCNN architectures. 

\begin{figure}[h]
\centering
\includegraphics[width=0.60\linewidth]{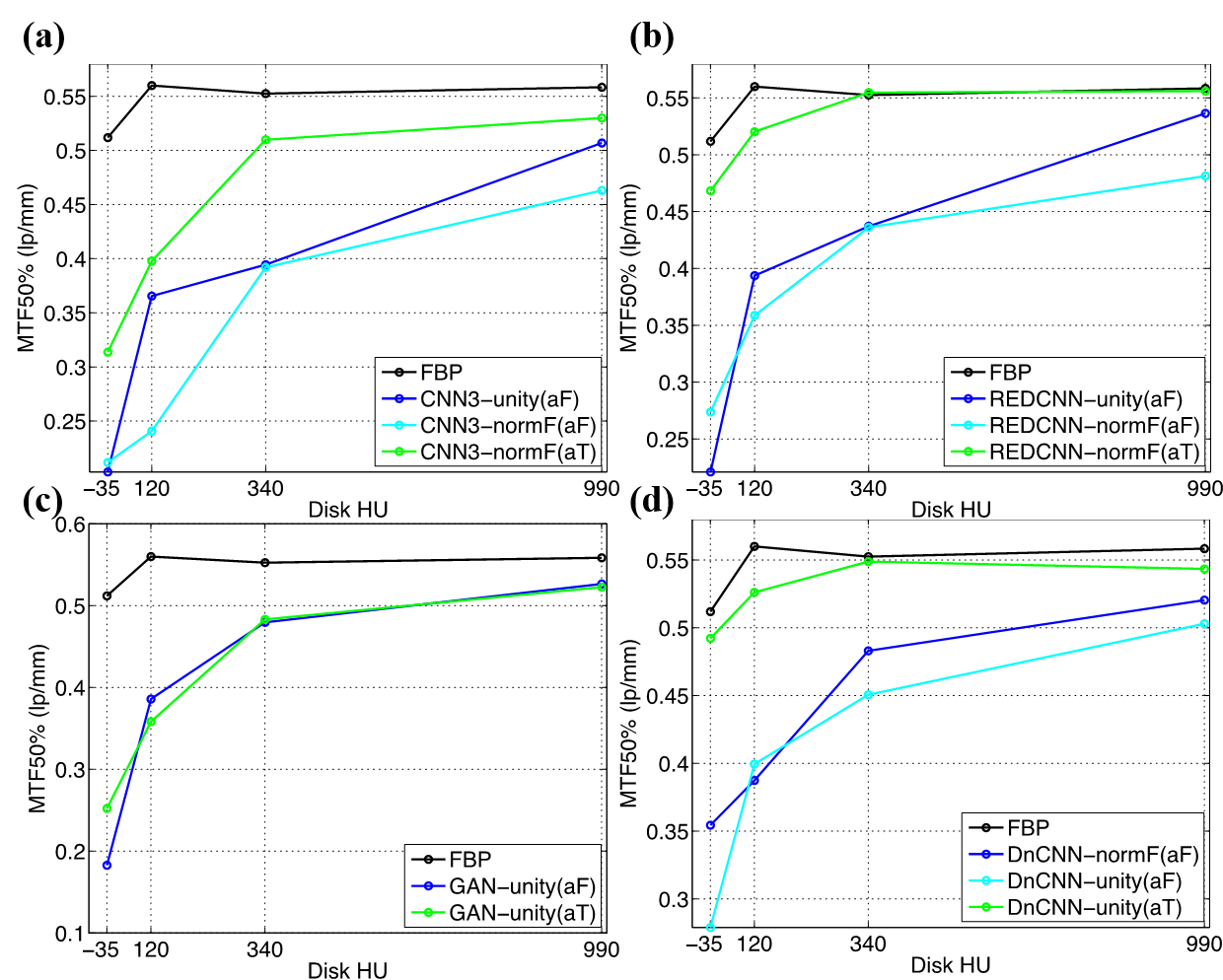}
\caption{MTF$50\%$ plots of (a) CNN3 (b) REDCNN (c) GAN and (d) DnCNN on the CAPHAN$600$ reconstructed using the FBP sharp kernel. Each plot here illustrates the network’s MTF$50\%$ performance trained making use of different normalization types and augmentations.}
\label{img:mtf_crgd}
\end{figure}
\subsubsection{MTF}
\label{sec:mtf}
\kjmodified{From the MTF viewpoint, we see that the MTF$50\%$ of the unity-normalized \& non-augmented (\textit{unity}(\textit{aF})) based learning is better than the rescaled \& non-augmented (\textit{normF}(\textit{aF})) based learning for the CNN3 and REDCNN (figs.$\ $\ref{img:mtf_crgd}(a-b)), \fsmodified{thereby} showing that the global metric-based learning (in section \ref{sec:glob_result}) supersedes the initial learning scheme bolstered by the CT numbers (in section \ref{sec:ct_numbers}) from the resolution viewpoint. However, both of these networks show significant improvement on MTF$50\%$ values once they are re-trained on \fsmodified{the augmented} dataset (\textit{normF}(\textit{aT})), \fsmodified{thus} adhering to the endpoint results from the section \ref{sec:ct_numbers}. }An analogous conclusion can be drawn for the GAN and DnCNN as illustrated in figs.$\ $\ref{img:mtf_crgd}(c-d).

\begin{figure}[h]
\centering
\includegraphics[width=1.00\linewidth]{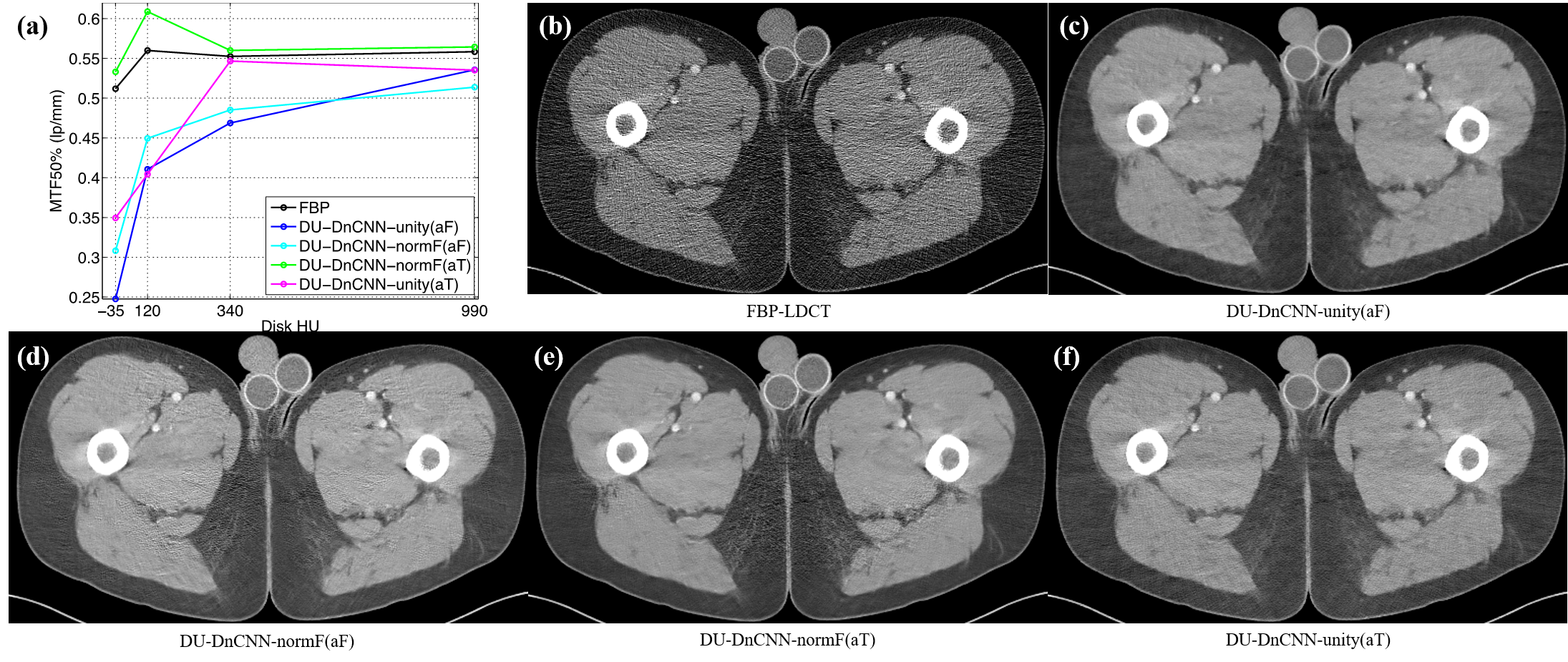}
\caption{MTF$50\%$ plot of (a) DU-DnCNN trained making use of different pre-processing schemes applied on the FBP reconstructed CAPHAN600. DU-DnCNN applied on (a) FBP-LDCT CT image trained making use of non-augmented \& unity normalized (b) unnormalized/rescaled \& non-augmented (c) unnormalized/rescaled \& augmented (d) unity normalized \& augmented pre-processing schemes. The display window consists of (W:$491$ L:$62$) HU.}
\label{img:unet_mtf_imgs}
\end{figure}

However, \fsmodified{the same} is not the case for the DU-DnCNN. First note that the best optimized scheme for the DU-DnCNN from the section \ref{sec:ct_numbers} - i.e.$\thinspace$\textit{normF}(\textit{aT}) - \rpzmodified{ surprisingly yields MTF$50\%$ values higher than that from the FBP} as illustrated by a green line in fig.$\ $\ref{img:unet_mtf_imgs}(a). This type of training even yields glass-like artifacts when applied to the LDCT test images. A representative example is depicted in fig.$\ $\ref{img:unet_mtf_imgs}(e) which is the denoised result when the DU-DnCNN-normF(aT) is applied to the FBP-LDCT image in fig.$\ $\ref{img:unet_mtf_imgs}(b). The DU-DnCNN-normF(aF) result, too, exhibits the glass-like discrepancies (fig.$\ $\ref{img:unet_mtf_imgs}(d)). Accordingly, we revert to \textit{unity} based pre-processing for the DU-DnCNN. Moreover, we proceed with the augmented-unity based DU-DnCNN (DU-DnCNN-unity(aT)) learning \kjmodified{because} its resolving power is higher than its counterpart trained without augmentation (DU-DnCNN-unity(aF)) as illustrated in fig.$\ $\ref{img:unet_mtf_imgs}(a).

\subsubsection{NPS}
\label{sec:nps}
\begin{figure}[h]
\centering
\includegraphics[width=0.60\linewidth]{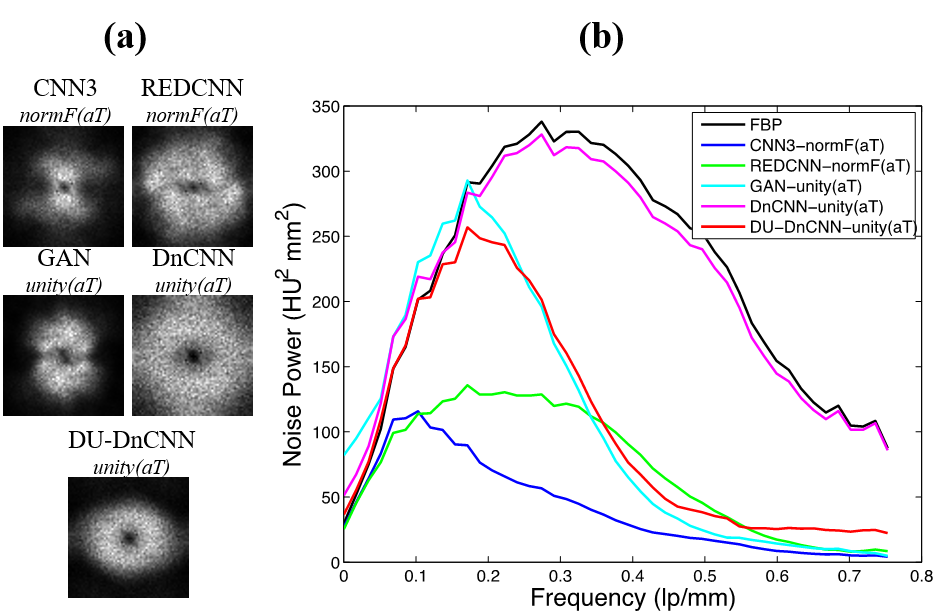}
\caption{(a) $2$D NPS images and (b) radial $1$D NPS curves of the networks (trained for the CT bench testing efficiency) applied on the noisy realizations of the reconstructed cylindrical phantom.}
\label{img:nps_4rm_ct_bench}
\end{figure}

The CNN3, REDCNN, GAN, DnCNN, and DU-DnCNN - that have been optimized to yield their respective best performances in terms of the HU accuracy and MTF - are further subjected to the NPS test as defined in section \ref{sec:glob_result}. \fsmodified{The} resulting 2D NPS images and 1D NPS curves are depicted in figs.$\ $\ref{img:nps_4rm_ct_bench}(a) and \ref{img:nps_4rm_ct_bench}(b) respectively. For the CNN3 and GAN, we notice that low to mid frequency bands (i.e.$\thinspace0.25$ to $0.4$ lp/mm) are no longer suppressed in the current 1D NPS curves (fig.$\ $\ref{img:nps_4rm_ct_bench}(b)) as compared to their counterparts from the previous global metric based learning (fig.$\ $\ref{img:nps_mtf_4rm_glob}(d)). Likewise, frequency bands along $0.45$ to $0.55$ lp/mm are more suppressed in the fig.$\ $\ref{img:nps_mtf_4rm_glob}(d) than in the fig.$\ $\ref{img:nps_4rm_ct_bench}(b) for the REDCNN and DU-DnCNN complimentary pairs. Most importantly, the doughnut shaped structure of the 2D NPS image resulting from the FBP-based reconstruction (fig.$\ $\ref{img:nps_mtf_4rm_glob}(c) top) is now replicated more appropriately by the five networks than before when they were optimized solely on the basis of the global metrics (fig.$\ $\ref{img:nps_4rm_ct_bench}(a) vs fig.$\ $\ref{img:nps_mtf_4rm_glob}(c)). It remains to be seen as to what degree or in what specific forms, these gains in the NPS values translate within a given CT image. We will \kjmodified{explore further} in the next subsection. 

\subsection{Difference image based analysis}
\label{sec:diff_img}

\begin{figure}[h]
\centering
\includegraphics[width=0.40\linewidth]{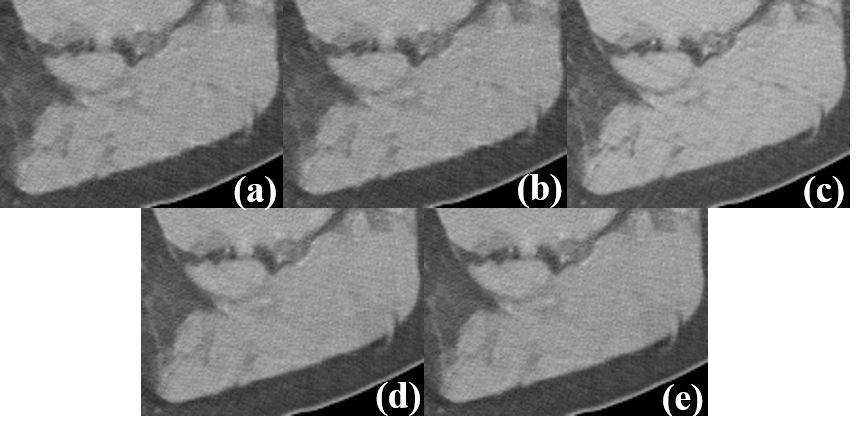}
\caption{ Denoised results from the (a) CNN3 [$35.45\ |\ 0.93$] (b) REDCNN [$35.21\ |\ 0.94$] (c) GAN [$33.44\ |\ 0.94$] (d) DnCNN [$35.49\ |\ 0.94$] (e) DU-DnCNN [$35.9\ |\ 0.90$] applied to the LDCT in fig.$\ $\ref{img:glob_abdomin_crop_view}(c). The display window consists of (W:$491$ L:$62$) HU.}
\label{img:ct_abdomin_crop_view}
\end{figure}

We begin our image analysis by performing direct comparisons between ROIs deduced by the five networks trained to yield best values from the global metric (i.e.$\thinspace$the SSIM, RMSE, PSNR) viewpoint to that trained to achieve a higher efficiency in terms of the CT bench tests. In particular, we apply the CNN3-normF(aT), REDCNN-normF(aT), GAN-unit(aT), DnDNN-unity(aT), and DU-DnCNN-unity(aT) trained weights on fig.$\ $\ref{img:glob_abdomin_crop_view}(c) to produce images depicted in figures \ref{img:ct_abdomin_crop_view}(a) through \ref{img:ct_abdomin_crop_view}(e) respectively. These five ROIs are the corresponding counter parts to the figures in \ref{img:glob_abdomin_crop_view}(e) through \ref{img:glob_abdomin_crop_view}(i) (in order) that are the outputs of the DNNs tuned from the global metrics viewpoint.  

\begin{figure}[hbt!]
\centering
\includegraphics[width=0.65\linewidth]{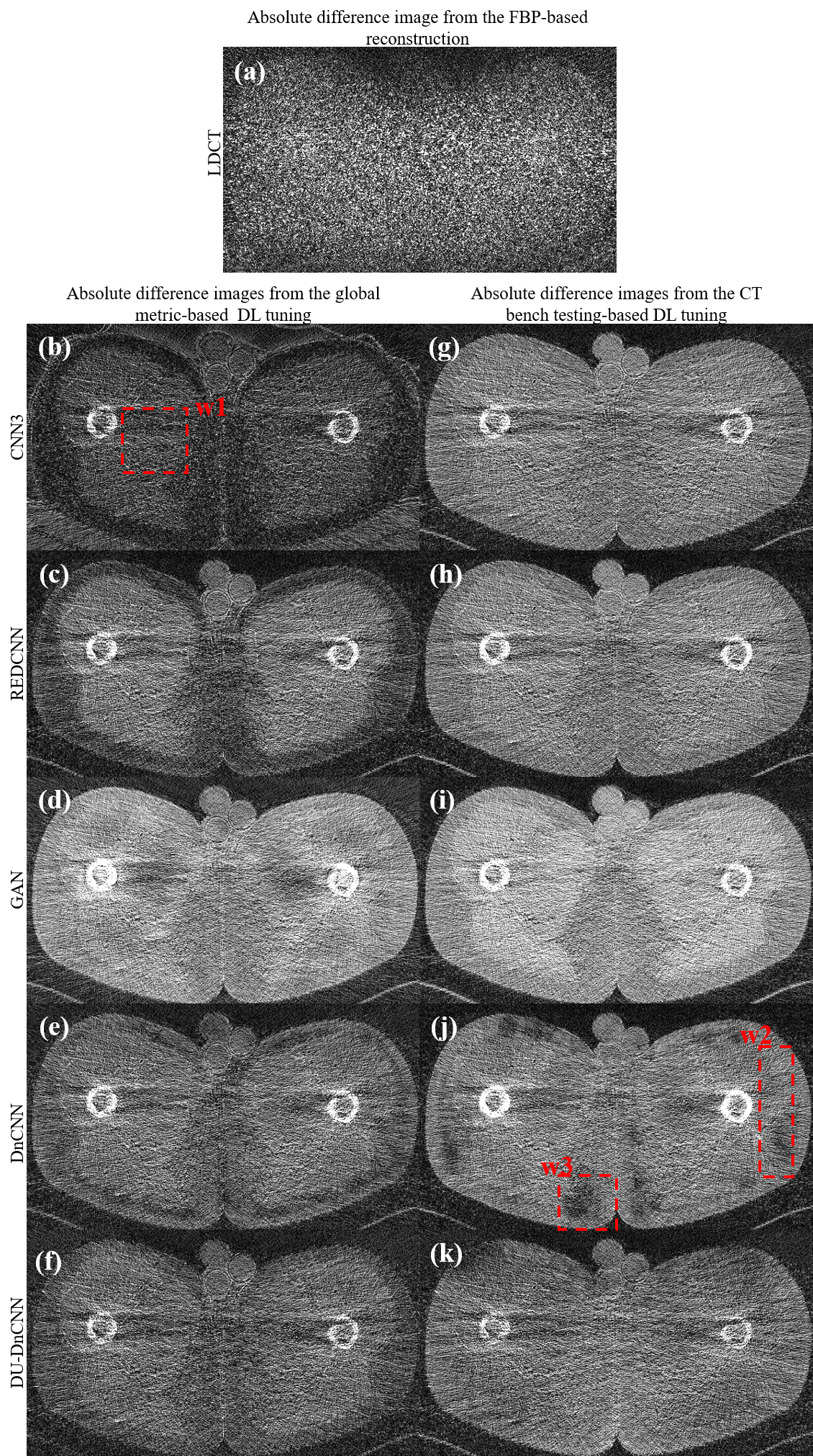}
\caption{Absolute difference image of (a) LDCT in relation to the NDCT. Other absolute difference plots between the NDCT and (b, g) CNN3 (c, h) REDCNN (d, i) GAN (e, j) DnCNN (f, k) DU-DnCNN. The plots in the left (b-f) are deduced from the DNNs trained to obtain the best performances from global metrics viewpoint and those in the right are re-trained to yield better performances from the CT bench testing methods. The display window consists of $[0, 122]$HU.}
\label{img:abs_diff_img}
\end{figure}

It is difficult to make any useful assertions on the state of the small anatomical features. For instance, if the \textit{M}-shape or the \fsmodified{features} inside the dotted circles are more discernible/consistent in figs.$\ $\ref{img:ct_abdomin_crop_view}(a-e) than in figs.$\ $\ref{img:glob_abdomin_crop_view}(e-i) when compared to their NDCT counterpart in the fig.$\ $\ref{img:glob_abdomin_crop_view}(b). Nonetheless, a clearer picture on the improvement of the overall denoising capacity of the networks – having gone through the refinements in section \ref{sec:ct_bench_result} – \rpzmodified{may} be drawn from the absolute difference image plots in fig.$\ $\ref{img:abs_diff_img}. We define absolute difference image for the LDCT as $|\mathbf{Y}–\mathbf{X}|$ and that for any given network as $|\mathbf{Y}–f_{\text{DL}}(\mathbf{X};\hat{\theta}_{\text{DL}})|$. $\mathbf{X}$ and $\mathbf{Y}$ represent the LDCT and NDCT image respectively and $f_{\text{DL}}$ and $\hat{\theta}_{\text{DL}}$ represent any given network architectures and its corresponding weights trained making use of the model (\ref{eq:general_loss_func}). \kjmodified{A quick glance of the fig.$\ $\ref{img:abs_diff_img} reveals that the plots in the figure's right column \fsmodified{are}, generally, more uniform than the plots in the figure's left column; \fsmodified{for instance,} absolute difference plots of the CNN3 in the figure's second row. Recall from the 1D NPS plot in the fig.$\ $\ref{img:nps_mtf_4rm_glob}(d) that the CNN3 exhibits the lowest NPS magnitude, out of all the five networks, throughout the plot’s frequency bands. This is particularly apparent along the bands above $0.3$ lp/mm.} Consequently, the absolute difference plot from the global metric based tuning of the CNN3 (fig.$\ $\ref{img:abs_diff_img}(b)) is plagued with \rpzmodified{edge-like structures} across the plot. This effect can be realized by comparing the two figures (figs.$\ $\ref{img:abs_diff_img}(b and f)) within the \fsmodified{dashed} window (w1). Likewise, the $1$D NPS curves in fig.$\ $\ref{img:nps_4rm_ct_bench}(b) of the DnCNN and DU-DnCNN resemble \kjmodified{more closely} to that of the FBP. More specifically, along the low-frequency bands from $0$ to $0.1$ lp/mm. Accordingly, the distinction between the outer and the inner regions (as highlighted by the \fsmodified{dashed} window (w$2$)) is the least for the DnCNN (in fig.$\ $\ref{img:abs_diff_img}(j)) and DU-DnCNNN (in fig.$\ $\ref{img:abs_diff_img}(k)) as compared to the rest of the plots in the fig.$\ $\ref{img:abs_diff_img}. Also, figs.$\ $\ref{img:abs_diff_img}(b-f), generally, exhibit sharp edges as compared to figs.$\ $\ref{img:abs_diff_img}(g-k). \fsmodified{These results suggest} that the improvement in the MTF$50\%$ values for all the five networks, having re-trained to optimize for CT bench testing, also reflects in terms of the gain in resolution of small anatomical structures. However, it should also be noted that the DnCNN’s absolute difference plot in fig.$\ $\ref{img:abs_diff_img}(j) exhibits patchy regions as illustrated by the doted window (w$3$). This suggests that the DnCNN’s architecture with $17$ hidden layers might have made it prone to overfitting in terms of the HU accuracy. Either a re-training with a lesser number of hidden layers or an analysis similar to the one with the DU-DnCNN in the section \ref{sec:mtf} for CT images from a different anatomical region(s) (such as the ones from lung or liver region) might be necessary to promote the DnCNN to learn a more generalizable solution.

\newaddition{
\section{CONCLUSIONS}
\label{sec:conclusions}
In this contribution, we considered an array of DNNs ranging from simple 3-layered to very deep 17-layered to encoder-decoder based to U-Net styled to generative adversarial based networks. Subsequently, we tuned them to yield the best performance in terms of the global metrics in the first approach and then to yield a better overall CT bench testing performance in the second approach. For all of them, the second approach seems to preserve the noise texture, resolutions of different contrasts and maintain the HU accuracy more appropriately than the former. These gains in the CT bench test performances reflect explicitly on CT images from the test set as indicated by the difference plots. In the near future, we seek to perform the CT numbers and difference plot based analyses on a wide range of anatomical features. In the long term, task-based assessments \cite{rpz_task_oriented} that reflect real life clinical settings will be considered to further generalize that the integration of CT bench test(s) - in combination with the PSNR, SSIM, RMSE - during the tuning phase yields the best diagnostic result for any given test set. 

Lastly, we note that the DNNs show promising results for the CT image denoising on the basis of the global image quality metrics. Also, the global metrics-based network tuning aids to minimize risks associated with overfitting and underfitting. However, it should also be noted that global metrics like the PSNR and the SSIM do not capture information on a DNN \rpzmodified{outputs' noise texture or its image resolution or its CT number consistency} which are of utmost importance to infer the DNN's denoising capacity from a diagnostic viewpoint. Therefore, as we charter on the road to unravel the mysteries that surround the DNNs and make attempts to increase their efficiency for medical imaging with new techniques, such as transfer learning, sinogram-based learning, hybrid learning with data consistency layers, gains on their performances should also be measured making use of the CT bench tests and subsequently, with task-based assessments. It is also useful to think of finding ways to feed information to the neural networks on their performances on CT bench test(s), alongside their performances on perceptual quality, as they learn their weights. For if we indeed seek that networks learn their optimal weights for medical imaging, they should yield the optimal result on the perceptual front as well as on the diagnostic front. \\
\\
\textbf{Disclaimer}: The mention of commercial products, their sources, or their use in connection with material reported herein is not to be construed as either an actual or implied endorsement of such products by the Department of Health and Human Services. This is a contribution of the US Food and Drug Administration and is not subject to copyright.\\
\\
\textbf{Acknowledgement}: We would like to thank CDRH Critical Path FY2020 for funding this project. 
}

\pagebreak


\end{document}